# High-resolution optical and acoustic remote sensing datasets of the Puck Lagoon, Southern Baltic

## Authors
Łukasz Janowski[1], Dimitrios Skarlatos[2], Panagiotis Agrafiotis[3], Paweł Tysiąc[4], Andrzej Pydyn[5], Mateusz Popek[5], Anna M. Kotarba-Morley[6], Gottfried Mandlburger[7], Łukasz Gajewski[8], Mateusz Kołakowski[8], Alexandra Papadaki[9], Juliusz Gajewski[1]

## Affiliations
1. Maritime Institute, Gdynia Maritime University, Długi Targ 41/42, 80-830 Gdańsk, Poland
2. Department of Civil Engineering and Geomatics, Cyprus University of Technology, Saripolou Street 2-6, Limassol 3036, Cyprus
3. Faculty of Electrical Engineering and Computer Science, Technische Universität Berlin, Einsteinufer 17, 10587 Berlin, Germany
4. Faculty of Civil and Environmental Engineering, Gdańsk University of Technology, Gabriela Narutowicza 11/12, 80-233 Gdańsk, Poland
5. Centre for Underwater Archaeology, Nicolaus Copernicus University in Toruń, Szosa Bydgoska 44/48, 87-100 Toruń, Poland
6. School of Humanities, University of Adelaide, Napier Building, North Terrace, 5000 Adelaide, South Australia, Australia
7. Technical University Wien, Department of Geodesy & Geoinformation, Wiedner Hauptstr. 8-10, 1040 Vienna, Austria
8. MEWO SA, Starogardzka 16, 83-010 Straszyn, Poland
9. School of Rural and Surveying Engineering National Technical University of Athens, 9 Iroon Polytechneiou Str. - 157 80, Zographou, Athens, Greece

corresponding author(s): Lukasz Janowski (ljanowski@im.umg.edu.pl)

## Abstract
The very shallow marine basin of Puck Lagoon in the southern Baltic Sea, on the Northern coast of Poland, hosts valuable benthic habitats and cultural heritage sites. These include, among others, protected *Zostera marina* meadows, one of the Baltic's major medieval harbours, a ship graveyard, and likely other submerged features that are yet to be discovered. Prior to this project, no comprehensive high-resolution remote sensing data were available for this area. This article describes the first Digital Elevation Models (DEMs) derived from a combination of airborne bathymetric LiDAR, multibeam echosounder, airborne photogrammetry and satellite imagery. These datasets also include multibeam echosounder backscatter and LiDAR intensity, allowing determination of the character and properties of the seafloor. Combined, these datasets are a vital resource for assessing and understanding seafloor morphology, benthic habitats, cultural heritage, and submerged landscapes. Given the significance of Puck Lagoon's hydrographical, ecological, geological, and archaeological environs, the high-resolution bathymetry, acquired by our project, can provide the foundation for sustainable management and informed decision-making for this area of interest.

## Background & Summary
Shallow water environments located in coastal zones are one of the most productive and valuable ecosystems on Earth contributing to nutrient cycling, carbon sequestration, and supporting a wide variety of marine species, including economically important ones. The tideless area of the Puck Lagoon is located on the Polish coast of the Baltic Sea[1] (Figure 1), in the eastern part of Puck Bay and the Gulf of Gdansk. It is considered to be the most valuable biodiversity hotspot on the Polish coast. It is separated from the open sea by the Hel Peninsula at the north-west and from the Puck Bay by the partly submerged Seagull Sandbar at the south-west (Figure 1A). The area covers 102.69 km$^2$ with average depth of 3.13 m and maximum depth at 9.4 m [2]. The shallowest parts of the Puck Lagoon (down to 2 m) cover approximately 30% of the area. The whole area is included in the Coastal Landscape Park and Natura 2000 PLB220005 and PLH220032 sites under the 'birds directive' and 'habitat directive'[3].

During the last few decades, the area was subjected to intense anthropogenic pressures due to pollution and nutrification, resulting in a considerable loss of benthic habitats and species, especially *phytobenthos*[4,5]. The Bay of Puck, nestled within the Baltic Sea, is particularly susceptible to the process of eutrophication, a consequence of nutrient influx from surrounding agricultural lands[6]. This nutrient-rich runoff, primarily composed of nitrogen and phosphorus compounds, is transported to the Bay of Puck via the network of rivers and streams that originate from these agricultural fields[7]. Although the northern parts of the Lagoon were dredged in five sites to support beach restoration in the seaside part of Hel Peninsula between 1989–1996[8], the ecological biodiversity of this area is still high. This is indicated by the occurrence of around 25 species of macroalgae, eight species of vascular plants, and over 30 species of benthic crustaceans and molluscs[9]. The Puck lagoon also has substantial archaeological potential, as shown by environmental research and previous archaeological discoveries[10]. Due to the lack of proper identification, and hence protection of this water area, the significant underwater cultural heritage is exposed to many threats.

Until now, the Puck Lagoon has lacked precise, high-resolution bathymetric data and backscatter intensity measurements, which significantly hindered conservation efforts and endangered underwater cultural heritage. Before this work, the only existing bathymetry layer for this area was generated through interpolation between Singlebeam Echosounder (SBES) measurements taken at 25 m intervals (Figure 1C). These measurements were obtained during June 13 – September 25, 2012 fieldwork, by the Hydrographic Office of the Polish Navy. Notably, the most recent hydroacoustic characterisation of seabed habitats within the limited expanse of the Puck Lagoon was undertaken in 2003. The primary measuring devices used for this purpose were SBES and Side-scan Sonar[11].

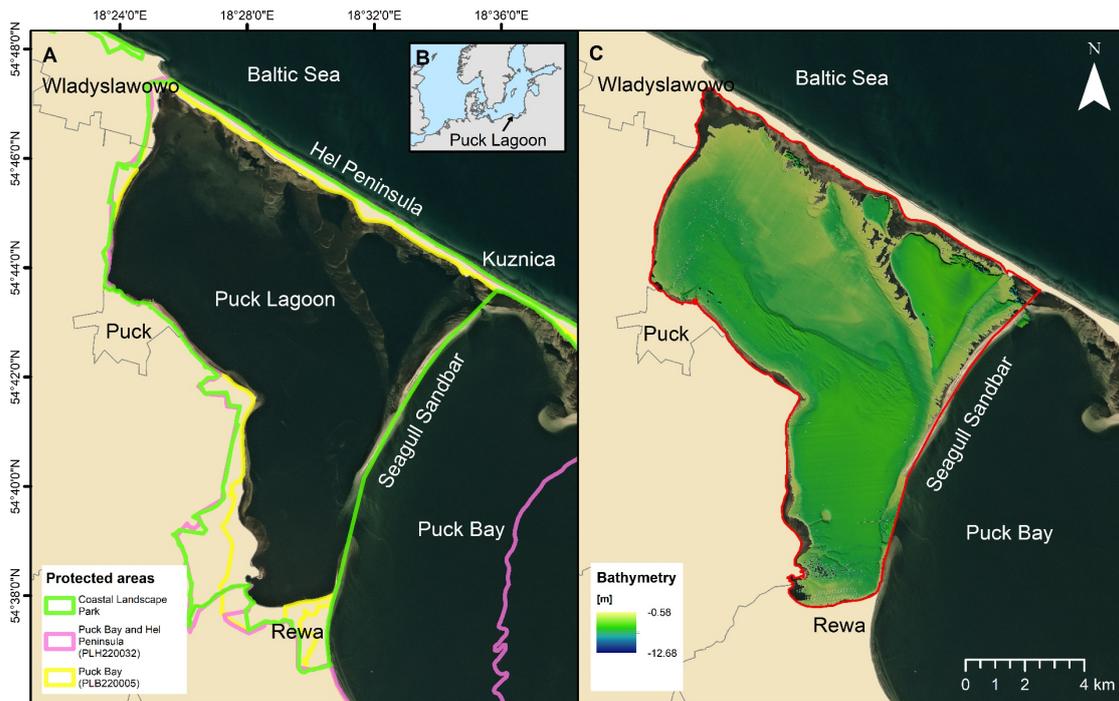

Figure 1. Overview of the study site and outlook to previous research. A – Location of the study site and its characteristic features. Satellite imagery provided by PlanetScope; B – Position of Puck Lagoon within the Baltic Sea; C – Bathymetry interpolated from SBES soundings recorded by the Hydrographic Office of the Polish Navy in 2012.

Shallow water environments, such as the Puck Lagoon, have not only ecological but also profound historical and cultural significance. The Baltic Sea has witnessed centuries of cultural exchange and the Puck Lagoon was a hub for maritime trade and transport in the medieval period. As home to one of the major medieval harbours in the Baltic, a bustling centre of commerce and trade, it welcomed ships from various corners of Europe[10]. Many of these ships sank and the seabed of Puck Lagoon is now an important submerged archaeological site not only for the remnants of the medieval harbour but also for the numerous wrecks that it contains. These shipwrecks hold invaluable insights into the maritime history of the region, spanning from medieval trading vessels to more recent naval (especially WWII) and merchant ships. Archaeological expeditions in this area have yielded artifacts that shed light on seafaring traditions, trade routes, and naval technologies of this important region of Northern Europe[12-14]. Environmental research conducted in the Puck Lagoon has consistently underscored the archaeological potential of the area[15]. However, these submerged cultural heritage sites are vulnerable to natural processes, anthropogenic pressures, and the passage of time. Proper identification, preservation, and understanding of this underwater cultural heritage are crucial to ensuring that they are protected and responsibly managed for future generations.

Investigation of the seabed habitats was undertaken in 2009 as part of a large national project[16]. This research identified two seabed types: soft bottom and post-dredging pits. Habitat types included: *Zostera marina* meadows, *Charophyceae* meadows, *Potamogeton spp.* and/or *Ruppia maritima* and/or *Zannichellia palistris* meadows. Bathymetry of the seabed was modelled based on archival soundings, additional artificial soundings, navigational charts, and satellite-based maps using the OASIS program (2009)[16]. Seabed habitats in this area were also studied under the "Zostera – Restitution of key elements of the inner Puck Bay ecosystem" project in 2010-2015 (www.zostera.pl). The project's results were generally focused on *Zostera marina's* spatial occurrence, one of the vulnerable, endangered species defined in Polish law. Benthic habitat maps in this study were created based on hydrodynamic models, satellite and ground-truth data[4]. One of the recent works presents categorisation of benthic habitats aligning with EUNIS 2019 classification system based on GIS analysis[17].

Apart from the bathymetry DEM, the current Multibeam Echosounder (MBES) and Airborne Laser Bathymetry (ALB) devices yield measurements of the intensity of the acoustic or laser signal from the seabed. These measurements, generally referred to as backscatter intensity, can serve as a proxy for various seabed characteristics. Whereas much more complicated than simple bathymetry, they offer a richer understanding of the seafloor. However, it is important to note that accurate determination of specific features such as sediment or vegetation types requires ground-truth information. Therefore, backscatter intensity is the basis of modern benthic habitat mapping discipline strongly related to the recent technical advancements of underwater remote sensing methodology. This multidisciplinary approach links knowledge from oceanography, underwater acoustics, ecology, sedimentology, geomorphology, statistics, geoinformation, geoengineering, geodesy and numerical modelling. Whereas the acquisition of MBES bathymetry is standardised for hydrographic purposes, understanding of intensity measurements of acoustic/laser signal from the seabed is much more complicated[18].

The primary aim of this study was to generate a comprehensive, high-resolution bathymetry dataset for the Puck Lagoon area of the Baltic Sea. This dataset, unprecedented in its detail, was compiled using a combination of bathymetric LiDAR, MBES, aerial photogrammetry, and Satellite-Derived Bathymetry[19]. The resulting dataset not only provides a detailed bathymetric map of the Puck Lagoon but also includes reflectance measurements.

The potential reuse value of this dataset is significant. It can serve as a valuable resource for addressing a wide range of ecological, geological, and archaeological questions. Furthermore, it can be utilized for sustainable management and conservation efforts in the Puck Lagoon area. The data generated by our research has significant long-term benefits and will support informed decision-making, aid in the preservation of biodiversity and cultural heritage, and contribute to the broader scientific knowledge of threatened coastal ecosystems in high-density tourist areas. By providing this dataset, we hope to fill a critical data gap and facilitate future research and conservation initiatives.

## Methods
### LiDAR data collection
The aerial survey of the Puck Lagoon was conducted by GISPRO SA under the supervision of the Maritime Institute, Gdynia Maritime University, during calm water and weather conditions from February 27 to March 2, 2022 (3 survey days). The precise weather conditions for the flight period and the research area were provided by the Polish Institute of Meteorology and Water Management (IMGW-PIB), the Zephr-HD model with resolution of 3 km for Europe (provided by Windguru and partners), and the EWAM wave forecast for Europe with resolution of 5 km (provided by the German weather service). The summary results for the three measuring days are presented in Table 1.

Table 1. Summary for weather conditions during flight measurements.

| Parameter \ day | 27.02 | 01.03 | 02.03 |
|---|---|---|---|
| **Temperature [°C]** | 1-3 | 0-5 | 0-5 |
| **Wind speed [km/h]** | 3-14 | 5-15 | 9-14 |
| **Precipitation [mm/h]** | 0 | 0 | 0 |
| **Significant wave height [m]** | 0.2-0.4 | <0.1 | <0.1 |
| **Wind waves [m]** | <0.1 | <0.1 | <0.1 |
| **Wave current [m/s]** | 0-0.3 | <0.1 | <0.1 |

The measurements were taken from an altitude of 600–700 meters by an SP-PRO Vulcanair P68 TC Observer plane equipped with a Topo/Applanix navigation system and GPS-IMU Type 57 recorder. The main measuring device used was a Riegl VQ-880-GII bathymetric laser scanner, which was integrated with two RGB and IR cameras. The manufacturer of the device declares maximum depth penetration of up to 1.5 Secchi depths. Bathymetric LiDAR measurements were planned to register at least 12 points per square meter. A standard quality assessment with the scientific laser scanning software OPALS confirmed a mean pulse density of 12 points/m$^2$ (median: 10 points/m$^2$) for the single lines. In the overlap area between adjacent flight lines the point density increased to approx. 25 points/m$^2$. In total, the entire lagoon was surveyed with 58 flight lines, each delivering data from an infrared laser channel for measuring water surface and alluvial topography and a green laser channel for bathymetry. The registered point cloud was generated in the PL-EVRF2007-NH vertical system and the UTM 34N projected coordinate system, based on the ETRS89 ellipsoid.

**LiDAR data processing**
Bathymetric LiDAR measurements were first analyzed in terms of trajectory alignment with the aircraft traverses. The trajectory alignment procedure was conducted with respect to a fundamental reference Global Navigation Satellite Service (GNSS) station that was situated approximately 20 kilometers from the furthest measurement site. To ensure the robustness of our airborne data, we scrutinized trajectory deviation statistics for any significant deviations. This rigorous approach to trajectory alignment and reference point adherence formed the foundation of our data processing workflow, ultimately enhancing the accuracy and reliability of our high-resolution datasets.

Refraction correction, including refraction of radiation beams and reduced beam propagation speed in water was performed in RiHydro software. To execute the refractive correction process effectively, a Water Surface Model (WSM) was meticulously constructed for each data acquisition series. This correction process served as a critical step in refining the bathymetric data, enhancing the accuracy of the final results, and mitigating the distortions caused by the refractive effects.

By incorporating these corrections, we aimed to provide reliable bathymetric data that can be used with confidence in various applications, contributing to the advancement of underwater mapping and research. In the data processing workflow, alignment was a critical step to ensure the accuracy and consistency of the extracted data. This alignment process was performed using a suite of Riegl software tools, namely RiPPROCESS,

SDCImport, and RiWORLD. The fundamental principle underlying this alignment process was the measurement of differences in the position of identical planes within the cross-sections of data rows. Systematic errors observed and measured during this phase were transformed into shift vectors, allowing for precise correction. RiProcess, a key component of our data alignment strategy, played a pivotal role in this endeavor. It was equipped to calculate and rectify a range of systematic errors, including shifts, drifts in all spatial directions (XYZ), variations in heading, as well as roll and pitch discrepancies. One crucial aspect of the data processing workflow involved the precise alignment of recorded lines of bathymetry airborne scans between each other and to reference points. In our processing procedure, we used automatic and the established points as a reference. To automatically identify and align planes within the dataset, we employed the "Plane Patch Filter" option, which was a feature available in the Riegl software[20]. This method offers an efficient means of detecting and aligning planes within the data. In the application of this method, one crucial aspect was the determination of normal vectors for the points comprising the planes within the dataset. In the case of a dense point cloud, two of the components were assigned to define the plane, while the third component specified the direction of the normal vector. The core concept behind this approach is rooted in the nearest neighbor method. There are two primary methods for implementing this concept. The first method involves employing the nearest neighbor approach for a predetermined number of points, while the second method utilizes all points within a specified distance from individual objects. In the context of this study, we opted for the latter approach, utilizing centroid constraining within distinct cubic regions. Therefore, the alignment was achieved through data optimization techniques and the Iterative Closest Point (ICP) algorithm[21,22].

After performing alignment procedures, we observed that the signal spectrum did not accurately mark targets at their correct heights leaving stepped traces between the measurement lines. In response to this issue, we fitted a nonlinear least squares solution to a surface that approximated the correct scanning results. This challenge led us to address several specific cases. First and foremost, we identified the areas where errors in point registration occurred, particularly in cases where the bottom classification alone was insufficient. Following this, we evaluated the level of detail in the bottom, a critical parameter indicating the quality of the acquired data. In the subsequent step, we needed to distinguish between points that accurately represented the bottom and those that did not. Once we had categorized these points accordingly, we constructed a surface based on the accurately registered points and aligned the erroneously positioned points with this surface. This approach proved to be pivotal in successfully enhancing the registration of the beam reflection amplitude. Contrary to attempts to determine the best-fitting model and incorporate structural information, directly deriving the beam reflection size from defined points in space yielded superior results.

Notably, many airborne scanning filtering methods[23,24] designated the last reflection as a ground point. However, in our exploration of air-water correction methods, we identified instances where points recorded as the last reflection did not accurately represent the ground. To address this, we implemented hierarchical filtering after dividing the point cloud into smaller, more manageable sections.

Our data filtering strategy was executed in a systematic manner, comprising the following key steps:

1. Division of Representative Area: We initiated the process by dividing the representative area into smaller, manageable sections. Within each of these sections, we carefully selected representative points. These representative points were identified through meticulous comparison with multibeam data, ensuring their correspondence with the actual bottom of the Puck Lagoon.
2. Approximation of Bottom Surface: Building upon the selected representative points, we proceeded to approximate the underlying bottom surface using triangulation techniques. This step allowed us to create an approximate representation of the Lagoon's seabed.
3. Hierarchical Point Rejection: To further refine the dataset, we implemented a hierarchical point rejection mechanism. This step involved identifying and rejecting points that were deemed too distant from the approximated bottom surface. By systematically eliminating outliers, we enhanced the overall data quality.

Once these filtration procedures were completed, we proceeded to implement vegetation filters locally to further refine and organize the data. This comprehensive filtration strategy ensured that the final dataset was not only accurate but also effectively processed for subsequent analysis and interpretation. After completing all the necessary data adaptation steps, we conducted a relative evaluation. This entailed assessing the quality of the airborne laser scanning data in relation to data obtained from the multibeam echosounder. By systematically addressing and correcting these errors, our data alignment process ensured that the final datasets are not only accurately aligned but also consistent, enabling reliable downstream analysis and interpretation of the acquired data. The high level of accuracy demonstrated by this alignment process, as grounded in our eigenvalue distribution analysis and point cloud extraction detailed in the 'LiDAR Data Quality' chapter, indicates the method's potential for innovatively enhancing the alignment and overall precision of airborne laser scanning data. This novel approach represents a significant methodological contribution by the authors aimed at expediting data processing and reducing the time required for field measurements, especially when compared to traditional methods employing total stations.

As an outcome of our data processing efforts, we presented the results in the form of reflection amplitudes displayed in grayscale (Figure 2B). Intensity values can be utilized to assess data quality through visual inspection. Anomalies or outliers in intensity may indicate data artifacts or errors in the scanning process. This can help improve data preprocessing and quality control procedures.

The processed LiDAR point clouds were used to generate bathymetry (Figure 2A) and intensity surface grids in GlobalMapper software. We used the Binning gridding method with manually specified grid spacing to 0.2 meters and default "no data" distance criteria. All grids were exported to GeoTiff data format. Grids were saved in PL-EVRF2007-NH vertical system, UTM34N projected coordinate system, based on ETRS89 ellipsoid.

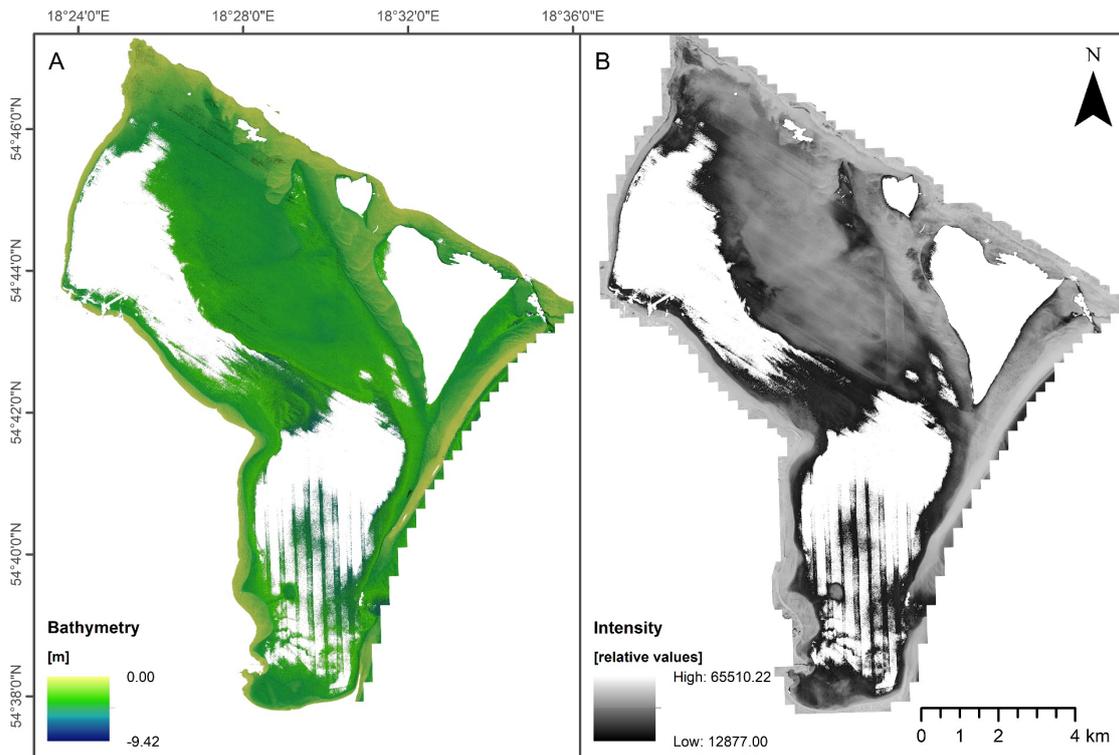

Figure 2. Bathymetric LiDAR datasets acquired in this study. A – Bathymetry grid; B – Laser intensity grid covering the same area.

## Aerial photogrammetry data acquisition and processing

Aerial photographs were acquired by GISPRO SA as an additional dataset to laser measurements during the same flights (described in the previous section). The aerial photographs were planned to ensure 80% longitudinal and 40% transverse coverage, with a terrain pixel size of approximately 8 cm and RGB and NIR colour composition.

## Refraction correction and Orthoimage generation

Although aerial image-based bathymetric mapping can provide both water depth and visual information, water refraction poses significant challenges for accurate depth estimation[25,26]. In order to tackle this challenge in this dataset, we implemented a state of the art image correction methodology[27,28], which first exploited recent machine learning procedures that recover depth from image-based dense point clouds[25,26] and then corrected the refraction effects on the original imaging dataset[27]. This way, the structure from motion (SfM) and multi-view stereo (MVS) processing pipelines were executed finally on a refraction-free set of aerial datasets, resulting in highly accurate bathymetric maps and respectively high quality and accurate orthoimagery.

Following the aerial image data collection and Ground Control Points (GCPs) measurement, an initial SfM-MVS was executed in order to obtain the required data for applying the proposed refraction correction methodology (i.e., the interior and exterior orientation of the cameras and the initial dense point cloud). For the SfM-MVS step, a specific software implementation did not affect the quality of the results, and they could be produced in a similar way using any commercial or open-source automated photogrammetric software, without employing any water refraction compensation. For the approach presented here, the Agisoft Metashape commercial software was used.

The resulting initial dense point cloud with systematic depth underestimation due to the refraction effect was corrected by employing the recently developed DepthLearn[26] solution for the seabed points. For DepthLearn, a Support Vector Regression model already trained on synthetic data[28] was used. This model enabled us to correct the effects of refraction on the point clouds of the submerged areas ignoring additional error sources not related to refraction e.g. waves, point cloud noise due to ambiguities in matching caused by the water column or refraction effect.

Consequently, this corrected point cloud was used to create an updated (merged) DEM with recovered seabed bathymetry which was then used to differentially rectify[29] the initial aerial image dataset. A new SfM processing was finally executed based on the refraction-free imaging dataset in order to update the interior and exterior orientation of the cameras, before orthomosaicing. During the last stages, texture and orthoimages can be generated based on the merged DEM generated by the corrected dense point clouds using DepthLearn and the initial dry-land points. Also the corrected dense point clouds and DEM can be exported to be used as bathymetry source (Figure 3).

Similarly like in the LiDAR datasets, we generated the unified grids from photo point clouds in GlobalMapper software with the same settings. All grids were exported to GeoTiff data format in PL-EVRF2007-NH vertical system, UTM34N projected coordinate system, based on ETRS89 ellipsoid.

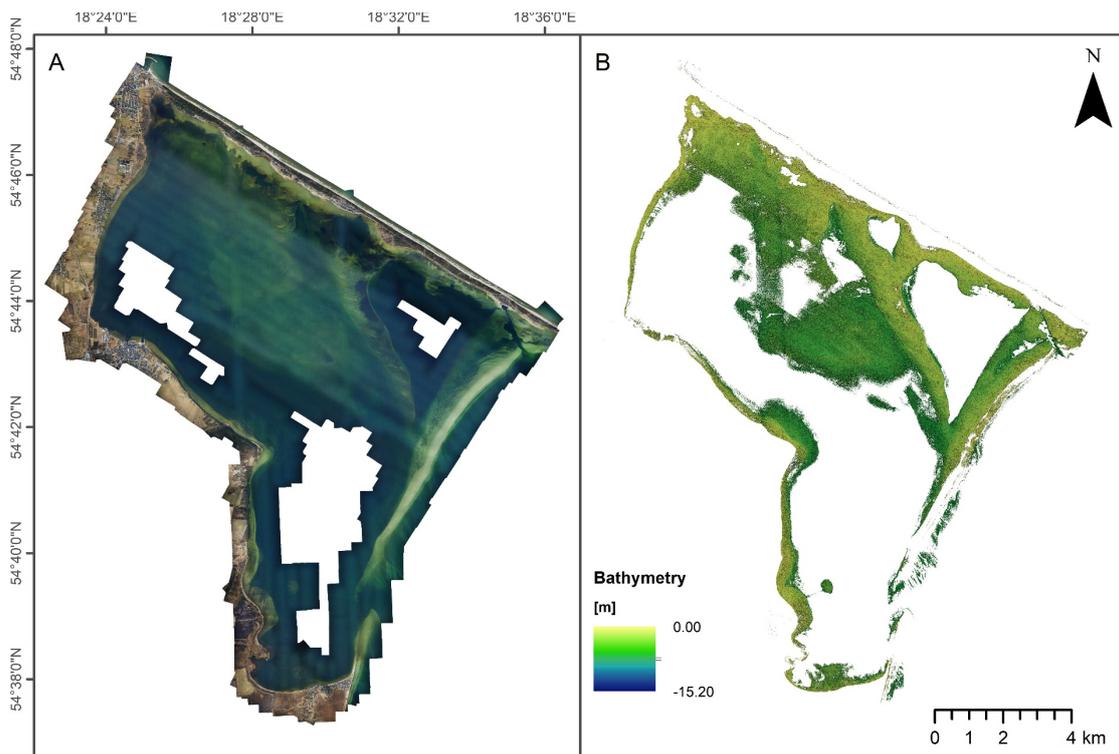

Figure 3. Aerial photogrammetry dataset acquired in this study. A – Orthophoto map for areas with measured terrain / bathymetry; B – Bathymetry grid generated with SfM technique from aerial photogrammetry.

**Multibeam data collection**
Multibeam echosounder (MBES) datasets were acquired during a three-month survey from March 22 to June 22, 2022, with a total of 21 surveying days. The main measuring device used was a Teledyne Reson T50-P or T20-P (from 23 May 2022), which was mounted on a pole onboard the IMOROS 2 or IMOROS 3 survey units. Both MBESs had the same settings throughout the survey, with the number of beams being 1024 in T50-P and 999 in T20-P (Table 2).

Table 2. Summary of MBESs settings.

| Feature \ MBES | T50-P | T20-P |
| --- | --- | --- |
| No. of beams | 1024 | 999 |
| Freq. [kHz] | 420 | 420 |
| Pulse length [µs] | 30 | 30 |
| Absorption [dB/km] | 70 | 70 |
| Spreading [dB] | 35 | 35 |
| Swath angle [°] | 130-140 | 130-140 |
| Power [dB] | 220 | 220 |
| Gain [dB] | 0 | 0 |

In addition to MBESs, the measurement and positioning system contained duplicated GPS receivers Trimble SPS851 and Trimble BX982, iXBlue Hydrins Inertial Navigation System, Reson SVP70 Sound Velocity Probe and Reson SVP15 Sound Velocity Profiler. All measurements were carried out with RTK corrections available for the whole area of interest. Multibeam echosounder recordings were collected using QINSy 8.18 software.

To ensure high-quality measurements, the MBES sensors were calibrated for time, pitch, roll, and yaw offsets and regularly checked. The sound velocity speed was measured with SVP15 at least every 6 hours, for each change of environmental conditions and always before and after every measurement session. The MBES survey was carried out to ensure at least 20% overlap between ship track lines, the measuring density not less than five points for a 1 m grid, and a reasonably constant speed (mean speed of 2-2.5m/s). The MBES survey was planned to cover the deepest areas within the study site (Kuźniza Basin, post-dredging pits, harbor entrance), as well as the shallower area with the potential archaeological significance (Figure 4A). MBES measurements were recorded in the PL-EVRF2007-NH vertical system and PUWG1992 projected coordinate system, based on the ETRS89 ellipsoid.

**Bathymetry data processing**

Raw bathymetry files from MBESs were processed in Beamworx Autoclean software. Data processing included application of Surface Spline filter to automatically remove outlier soundings from the measurements. The other filter used for bathymetry data processing was Shift Pings to Neighbors. This filter works only when there exists at least some overlap between neighbors and it shifts the survey lines to its neighbors by using a best fit algorithm. The datasets were manually checked to remove any remaining erroneous soundings[30].

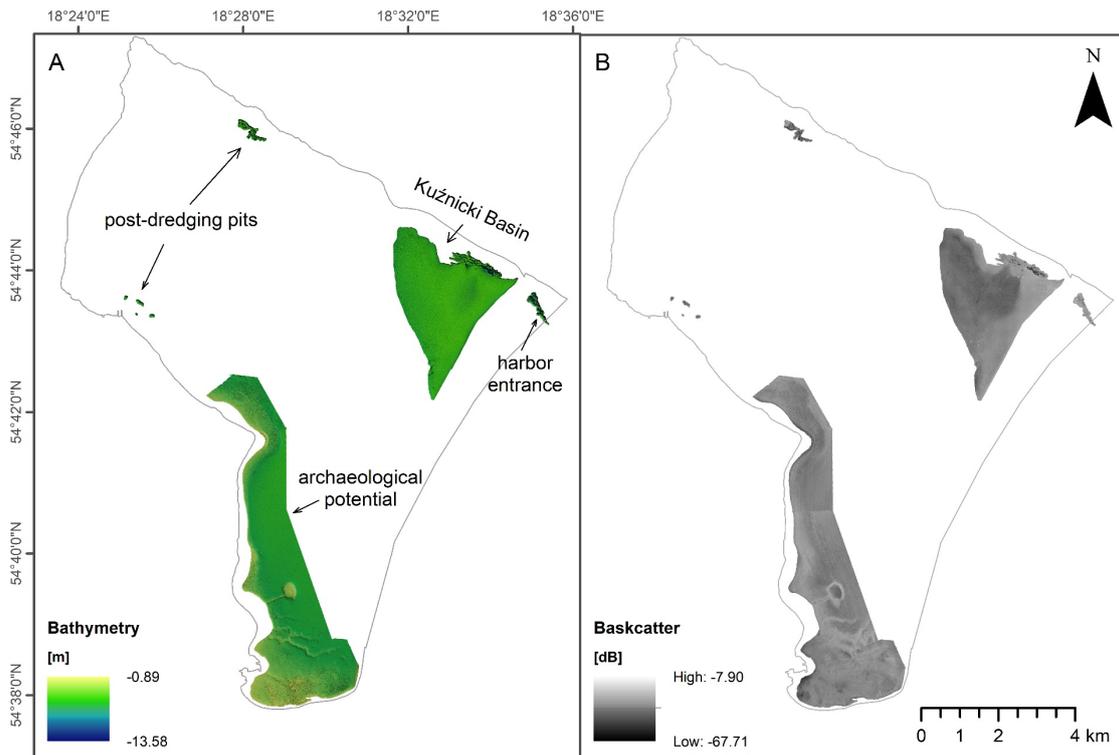

Figure 4. Multibeam echosounder dataset acquired in this study. A – Bathymetry grids; B – Grids of backscatter intensity of acoustic signal.

MBES Backscatter (Figure 4B) data processing was performed in Fledermaus Geocoder Toolbox (FMGT) software. The cleaned MBES soundings, as well as the bathymetric surface, were imported to an FMGT project. Data processing included application of Angle Varying Gain (AVG) filter to compensate backscatter measurements for angular variations[31]. We used the following settings of AVG filter: 300 pings sliding window and "flat" algorithm. All radiometric and geometric corrections were automatically applied using the default settings.

Finally, all bathymetry and backscatter datasets were exported to surface grids in GeoTiff data format with 0.2 m resolution. To ensure unification between all datasets, grids were saved in PL-EVRF2007-NH vertical system, UTM34N projected coordinate system, based on ETRS89 ellipsoid.

**Satellite-derived bathymetry data acquisition and processing**
Satellite-derived bathymetry (SDB) was obtained based on a 4-Band SPOT 6 satellite image acquired on April 19, 2021 (Figure 5A). The image was provided by Apollo Mapping LLC, USA in primary data format, without tiling, 16-bit pixel depth, and reflectance radiometric processing (without pansharpening and orthorectification). The SPOT satellite image allows reaching a panchromatic resolution of 1.5 m and an RGB pixel resolution of 6 m. The used image has dimensions of 3070 x 2541 pixels and a pixel size of 5.218 meters x 8.311 meters. The coordinate system of the predicted depths is the EPSG:25834 - ETRS89 / UTM zone 34N while the vertical Datum is the PL-EVRF2007-NH.

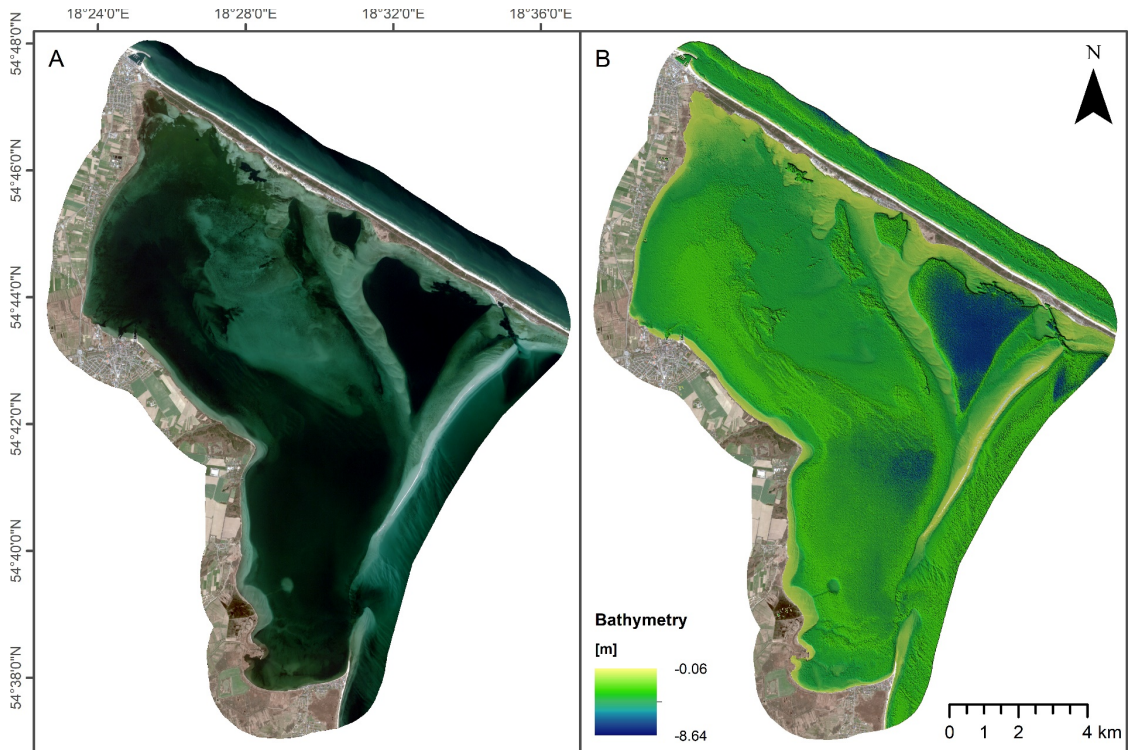

Figure 5. Satellite-derived bathymetry approach utilised in this study. A – The SPOT6 MS image used; B – Estimated Satellite Derived Bathymetry superimposed to the used SPOT6 image (only dry land visible in RGB).

Empirical SDB methods, like machine learning methods, require certain bands in the visible wavelengths, with blue and green being the most widely used as independent variables, and a set of known in situ depths (the dependent variable) as the only inputs in simple or complex models trained to deliver bathymetry estimations in a given area. For delivering SDB in the optically clear and shallow waters of this dataset, a Random Forest approach was implemented, exploiting the blue (0.45–0.52 µm), green (0.53–060 µm) and red (0.62–0.69 µm) SPOT6 bands. As a pre-processing step, dry land was masked in order not to affect training. A random forest is an ensemble learning method for classification, regression and other tasks that fits a number of decision tree classifiers on various sub-samples of the dataset and uses averaging to improve the predictive accuracy and control over-fitting. For classification tasks, the output of the random forest is the class selected by most trees. For regression tasks such as the SDB one, the mean or average prediction of the individual trees is returned[32]. In the performed approach 300 trees were used and Mean Standard Error as a criterion for accuracy. The resultant SDB bathymetry has a pixel size of 5.218x8.311 meters. The value of -0.293746 m in certain areas was used to represent no data.

A very important feature of the performed SDB approach is the use of already available in situ bathymetric data to train the RF model. Usually, such data are collected by expensive state-of-the-art equipment such as airborne LiDAR or MBES systems. In order to cover the complete area to be mapped, a combination of ALB and SBES data was used. The use of the SBES was considered necessary since the ALB coverage was limited to the shallower parts of the lagoon. Similarly, MBES had limited partial coverage of the area. Towards that, ALB missing data were covered by an older SBES campaign performed in the area (the 2012 dataset covered by the Hydrographic Office of the Polish Navy mentioned in the Introduction). Predicted SDB is illustrated in Figure 5B in colorscale.

**Integrated bathymetry**

We combined bathymetry measurements from LiDAR and MBES, integrating them into a comprehensive Digital Elevation Model. This integration was executed in the SAGA GIS software, utilizing the Mosaicking procedure with specific parameters: feathering of overlapping areas, a blending distance of 100, blending boundary for valid data cells, and the regression option for match. Feathering works by estimating a weighted average to find a target value for overlapping cells, using a fifty-fifty weighting when the boundary distance of the two grids is equal. The match option conducts a linear regression based on the values of all overlapping cells. This is done to align the values of the processed grid with those that have already been incorporated into the mosaic. The integrated bathymetry is provided in Figure 6.

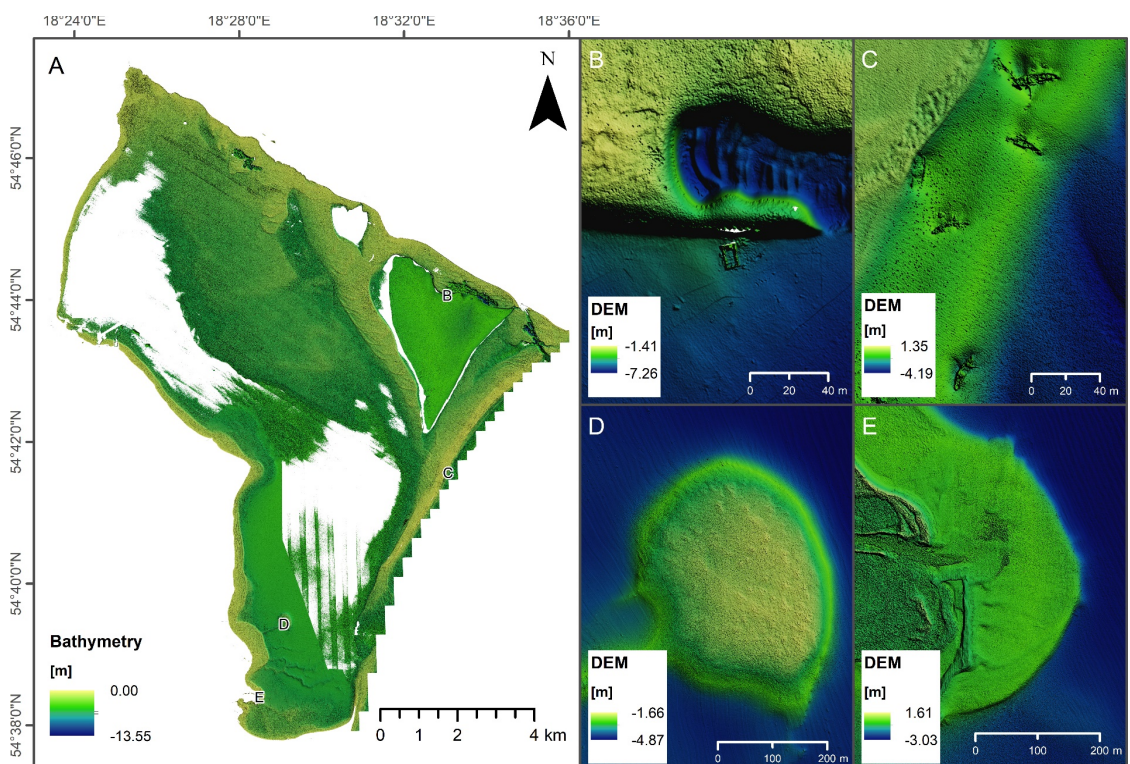

Figure 6. Overview on integrated bathymetry from MBES and LiDAR measurements. Vertical Exaggeration used: 5.0. A – Map on the integrated dataset. Capital letters on the map represent panel images provided on the right side; B – Part of dredging pit with the artificial structure remaining on the seafloor; C – Shipwrecks graveyard on the Seagull Shore; D – Oval shoal formerly considered to be a submerged settlement; E – Mouth of the Reda river flowing into Puck Lagoon.

**Data Records**

The remote sensing grids described in this paper can be assessed through the Marine Geoscience Data System (MGDS)[19]. All datasets are provided in GeoTiff raster format, and the files are named according to the data acquisition source.

The DEMs with a resolution of 0.2 m are available under the following names: MBES_DEM (from Multibeam Echosounder), LIDAR_DEM

(from Light Detection and Ranging), PHOTO_DEM (from Photogrammetry), and INTEGRATED_DEM (from Integrated MBES/LiDAR data). An additional DEM file, SDB (from Satellite-Derived Bathymetry), is also available but with a coarser resolution of approximately 5 m x 8 m.

The last three files represent seafloor/terrain properties at a 0.2 m resolution. They are named as follows: MBES_BSS, which represents the backscatter of the acoustic signal; LIDAR_INTENSITY, which represents the intensity of the laser signal; and ORTHOPHOTO, an orthorectified mosaic of aerial photographs of the research area.

## Technical Validation

In this section, we present a comprehensive approach for validation of each dataset. Since some of the methods are not yet frequently used for hydrographic approaches, we have dedicated much more attention to them compared to the well-established methods for bathymetry measurements (MBES). It is worth noting that all measurement campaigns were performed in favourable weather and environmental conditions that were strictly noted and checked during all surveys. It is extremely important not only for marine acoustic surveys, but particularly for airborne LiDAR or photogrammetry measurements. In fact, ideal environmental conditions for such measurements in Southern Baltic Sea may occur only for several days a year[33].

### LiDAR data quality

To accurately assess the obtained data, we commenced the evaluation process with the analysis of the trajectory alignment, which delineated the path of the aircraft. The fundamental principle underlying the approach of trajectory alignment was to acquire data at a single reference station. On average, the field-level details exhibited accuracies of approximately 3 cm, providing a robust foundation for alignment and subsequent analysis. Hence, it is essential to ascertain additional parameters that represent the geometric quality of satellites during the conducted measurement activities. These values are presented in Table 3.

Table 3. GNSS geometrical statistic values during flight mission.

| Statistics | Min | Max | Mean |
|---|---|---|---|
| **Baseline length [km]** | 13.09 | 44.82 | |
| **Number of GPS SV** | 6 | 9 | 8 |
| **Number of GLONASS SV** | 4 | 6 | 5 |
| **Number of QZSS SV** | 0 | 0 | 0 |
| **Number of BEIDOU SV** | 0 | 0 | 0 |
| **Number of GALILEO SV** | 0 | 0 | 0 |
| **Total number of SV** | 11 | 15 | 13 |
| **PDOP** | 1.28 | 2.12 | 1.47 |
| **QC Solution Gaps** | 0.00 | 0.00 | |
| **Solution Type** | Fixed | Float | No solution |
| **Epoch (sec)** | 1705.00 | 0.00 | 0.00 |
| **Percentage** | 100.00 | 0.00 | 0.00 |

Based on the aligned trajectory and its assessment, laser scanning points were extracted, followed by the relative combination of these extracted series into automatic planes and measured points.

Our evaluation of the acquired data encompassed the determination of height point errors (for the absolute model data). The alignment process was conducted employing the 'least squares' method, utilizing both manually and automatically established control points. Each data row featured several parameters considered variable for compensation: Roll [degrees], Pitch [degrees], Yaw (heading) [degrees], East [meters], North [meters], and Height [meters]. In total, there were 345 free parameters, with 40,000 planes serving as observations. Manual control points were acquired through the differential GNSS technique, with reference to the reference base station. Automatic control points were identified using an iterative algorithm that searched for planes within cubic regions utilizing the octree method. This method involved fitting a plane among points located within a cube, ensuring an error value not exceeding the user-defined threshold. Subsequently, the radius and angular deviation were set, serving as the maximum acceptable values for the compensation process. The alignment process was iterative, where each iteration introduced adjustments within the specified tolerance. A tolerance threshold of 0.0001 meters was employed. The evaluation of alignment quality was gauged through the standard deviation between points. The average standard deviation error for the scan lines of the whole flight was computed to be 0.0647 meters.

In the final phase of our technical test, we performed feature extraction based on covariance matrix. This matrix helps determine the directions of our data. The number of directions, or eigenvectors, matches the number of dimensions in our dataset. Since our point cloud data exists in three dimensions (X, Y, Z), each point in the dataset has three eigenvectors and three corresponding eigenvalues, which indicate the variances along each direction. These eigenvalues are essentially coefficients linked to the eigenvectors, providing insights into how much the data varies in specific directions. Computations were performed in CloudCompare software. In our study, we used the following tools for this purpose:

1. Roughness. It refers to local deviations from the plane. The feature is equal to the distance between the selected point to the fitted plane (minimum of 3 points is required to fit the plane). Different values of the feature can indicate, among other things, areas contaminated with noise, differences in registered points on the bottom, or local irregularities. The distribution of roughness values is presented in Figure 7A.

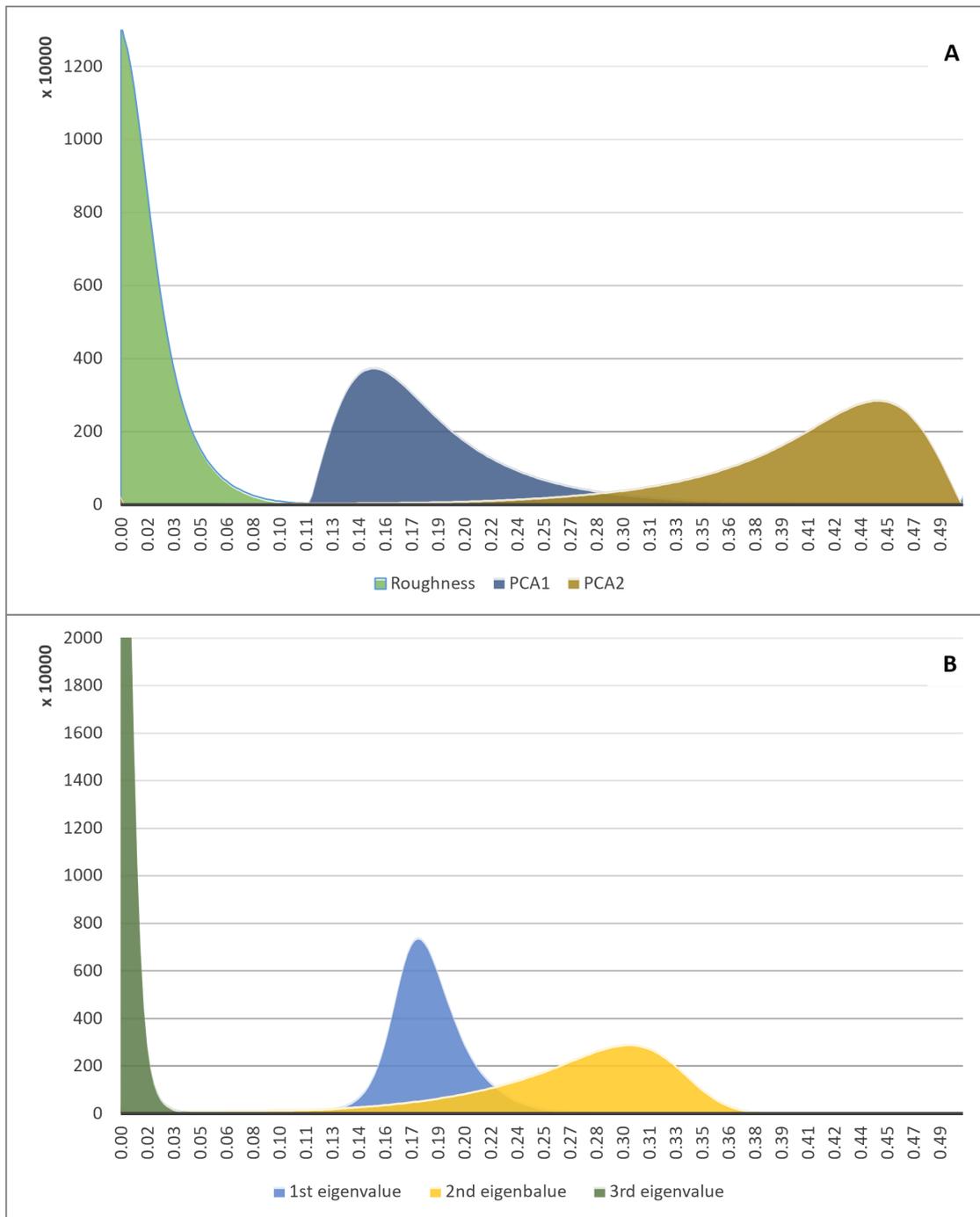

Figure 7. Plots representing technical validation of LiDAR dataset. Variables are marked with representative colours. A – distribution of roughness, PCA1 and PCA2 features; B – distribution of 1$^{st}$, 2$^{nd}$, and 3$^{rd}$ eigenvalues.

2. The eigenvalues of a covariance matrix provide insights into the patterns of variation and changes within the data's local structure. The first eigenvalue is associated with the first eigenvector of the covariance matrix for a given point, indicating the direction of the greatest variance. The second and third eigenvalues represent, respectively, the mean and the smallest covariance values, thus indicating the average and the least variable data directions. Significant local deviations in each of these directions within a specific object fragment may imply noticeable changes in its orientation. The results are presented on Figure 7B.
3. Principal Component Analysis (PCA). PCA is a technique used to reduce the dimensionality of datasets, minimizing data loss. It's based on eigenvectors and eigenvalues. For a three-dimensional dataset, there are three principal components. The first principal component (PCA1) captures the maximum variance in the dataset, while the second (PCA2) captures the second highest. Subsequent components contain less data and can be omitted to reduce dataset size with minor accuracy loss. In the software used, PCA1 and PCA2 are point clouds derived from dividing the first and second eigenvalues by the sum of all three, assessing the significance of data variation directions. Dividing an eigenvalue by the sum of all three gives its percentage contribution to total data variance, ranging from 0 to 1 due to normalization. A value close to 1 indicates a significant contribution, while a value closer to zero is less significant. If no principal component is dominant, their values will be in the middle range. Results of PCA analyses are shown in Figure 7A.

The quality of a LiDAR dataset was assessed by examining the spread, skewness, and peaks of the provided features (Figure 7). A high degree of skewness in the roughness or eigenvalues might suggest inconsistencies in the data collection process, whereas a wide spread in the PCA features could indicate a diverse and complex terrain.

The roughness (green) is highly skewed towards lower values. This skewness towards lower roughness values indicates that the majority of the LiDAR data points represent a relatively smooth bathymetry. In terms of data quality, this suggests that the LiDAR system was able to accurately capture the smoothness of the terrain, which is a positive indicator of data quality. PCA1 (blue) is moderately distributed but still skewed, suggesting a certain level of variability in the data. This variability in PCA1 could be indicative of the LiDAR system's ability to capture different terrain features, which is another positive aspect of data quality. PCA2 (yellow) appears almost normally distributed, indicating a balanced dataset in terms of this feature. A balanced distribution in PCA2 suggests that the LiDAR system was able to evenly capture the features of the terrain, further supporting the quality of the data.

In addition, the 1st eigenvalue (green) is highly skewed towards lower values, suggesting that for most points in the dataset, the direction of greatest variance is small. This could indicate that the points are closely packed along one direction, which might be the case if the LiDAR data represents a flat surface or linear feature. The 2nd and 3rd eigenvalues show more balanced distributions, indicating that these components contribute more evenly to the total variance. This suggests that the average and least variable directions in the data have a wider range of values. In terms of data quality, this balanced distribution in the 2nd and 3rd eigenvalues suggests that the LiDAR system was able to capture a variety of terrain features, which is a positive indicator of data quality.

In summary, the skewness, spread, and peaks of the features in the LiDAR dataset provide valuable insights into the quality of the data. The skewness in the roughness and 1st eigenvalue, the variability in PCA1, the balance in PCA2, and the even contribution of the 2nd and 3rd eigenvalues all suggest that the LiDAR system was able to accurately and comprehensively capture the terrain, which is indicative of high-quality LiDAR data.

## Aerial photogrammetry and refraction correction data quality

### Image alignment and dense image matching quality

Structure-from Motion and image overlap allowed to capture a total of 4,384 images from a flying altitude of 676 m, however, due to the poor bottom texture, only 3,473 of these images were aligned correctly. In total 1,649,970 tie points were found with 6,143,715 projections and a reprojection error equal to 1.19 pixels. The dense point cloud counted 236,775,975 points. Estimations of camera positions errors confirmed that larger errors are apparent in the outer region of the block and over areas with poor bottom texture. Regarding the camera location estimation, the average camera location error equals with 1.21 cm for the X axis, 1.29 cm for the Y axis, 2.39 cm for the Z axis resulting in an average total error of 2.97 cm. The same errors for the 22 GCPs used are calculated as follows: 1.15 cm for the X axis, 0.71 cm for Y axis, 0.69 cm for Z axis and 1.52 cm in total. The larger errors can be found in the northern part of the study area, and as expected, this negatively affected the image-based bathymetry estimated for this area.

During the SfM-MVS processing and refraction correction, it was decided to deliver orthoimages only for the areas were refraction corrected point cloud was available, rather than use hole filling tools etc. to deliver an orthoimage of the whole area, however, with limited accuracy in the areas with incorrect DEM.

### Refraction correction and final 3D point cloud quality

After performing SfM-MVS, the generated 3D dense point clouds and refraction corrections have been compared to the ALB data collected and described above. In this context, it is reported that before the refraction correction steps, by comparing to the ALB data, an RMSE of 0.956 m, a mean of 0.707 m, and a standard deviation of 0.643 m were calculated. However, after the refraction correction steps, the same metrics were formed as follows: RMSE equals to 0.420 m, mean to -0.0427 m, and standard deviation to 0.416 m. The majority of the points present distances to ALB around 0 m. However, there are still some points with much larger differences, being responsible for the RMSE and standard deviation value. Remaining errors are mainly due to noise in the point cloud because of the mainly homogenous bottom in the majority of the area as well as errors in the areas covered by the image-based point cloud neighbouring with the areas of "optically" deep waters. It is certain that the lack of proper texture in the bottom of the lake affected the SfM-MVS processing of the aerial data in a quite negative way. Also, errors are introduced due to issues in the GCPs of the northern part of the area.

### MBES data quality

According to the IHO standards, the quality of the MBES data was very high. Almost all of the MBES footprints (99.83%) met the IHO Special Order standard[34], which requires the following maximum parameters of Total Vertical Uncertainty at 95% confidence level: a = 0.25 m, b = 0.0075 m. The variable 'a' represents the systematic uncertainty that remains constant regardless of the depth. On the other hand, the variable 'b' represents the random uncertainty that varies with depth[34]. Only 0.09% of the footprints were rejected due to errors or outliers. The total number of accepted footprints was 2,576,203,403, while the total number of rejected footprints was 2,206,654. Table 4 shows more details about the validation statistics.

Table 4. Attribute statistics for MBES survey.

| Attribute | Mean | Minimum | Maximum |
|---|---|---|---|
| **Span** | 0.01 | 0.00 | 21.49 |
| **95% Confidence Level** | 0.01 | 0.00 | 17.36 |
| **Survey Accuracy** | -0.24 | -0.27 | 17.11 |

**SDB data quality**

The horizontal accuracy of SDB is a function of the spatial resolution of the satellite sensor used and the uncertainty contained in the in-situ bathymetric data used for training the Random Forests model. However, in Satellite Derived Bathymetry, the achieved vertical accuracy is crucial. This is heavily affected by the water column characteristics and seabed texture and habitat, meaning that areas with highly turbid waters, chlorophyll-a concentrations or dark colored sediments would deliver depths with larger uncertainties.

In the presented work, the estimated Satellite Derived Bathymetry as well as the efficiency of the Random Forest model were evaluated. For the validation approach, we used and provided pixel-based quality information on the reliability of the deliverables. To that end, unseen data to the model, i.e. the remaining 10% of the data not used for training, were used to calculate the RMSE, MAE and $R^2$. Specifically, the total size of the available data for training was 1,214,456 points of which the 1,093,010 points (90.0 % of used sample) were used for training while the remaining 121,446 points (10.0 % of used sample) for testing the model and calculating the RMSE which equals to 0.498 m, the MAE equal to 0.297 m and the $R^2$ equal to 0.917.

In large parts of the Puck Lagoon, SDB achieved a very high accuracy. However, there are still areas of erroneous predictions, attributed mainly to the lack of visibility of the bottom, i.e., optically deep waters, where the method is unable to deliver depths, since bottom reflectance is not captured by the satellite sensor. Whereas the majority of the vertical differences present values close to 0 m, there were some outliers up to 6 m of difference.

**Integrated DEM data quality**

The quality of the integrated dataset was assessed through a cross-validation process between the integrated dataset and the MBES/ALB datasets. In the reference area, which represents the densest coverage between MBES and ALB (indicated by the dashed line in Figure 8), we generated regression scatterplots to track their correlation with the integrated DEM dataset. The R-squared results demonstrated an extremely high correlation between the MBES and Integrated dataset (R2=0.9998; Figure 8A), as well as between the Lidar and Integrated dataset (R2=0.9992; Figure 8B).

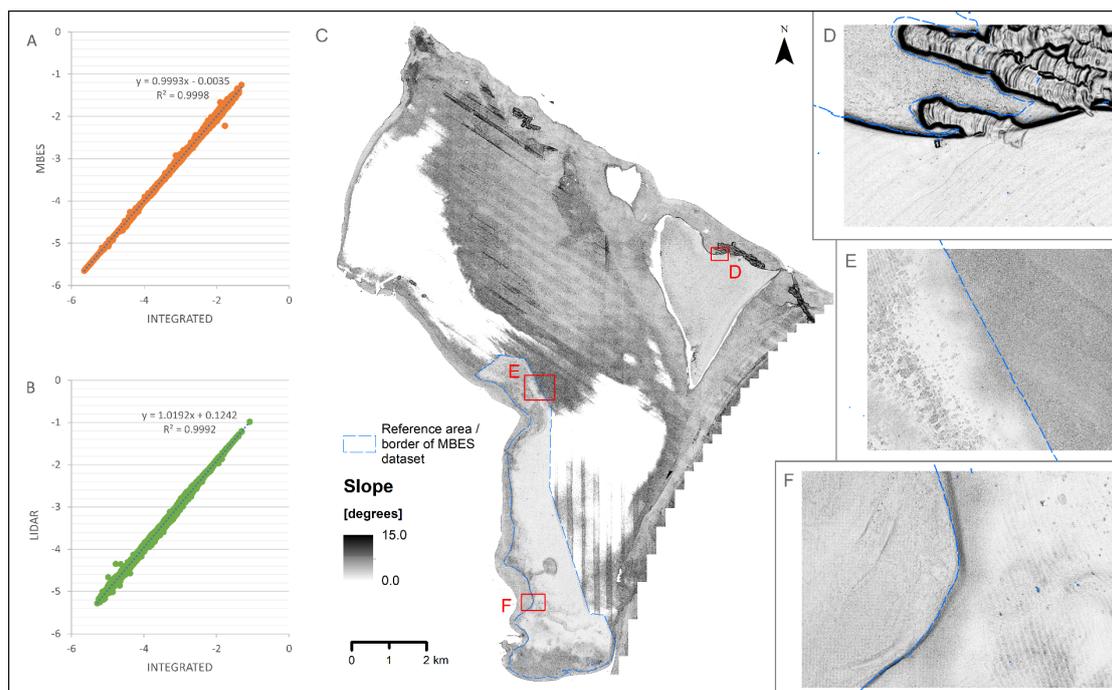

Figure 8. Results of cross-validation and spatial representation of the slope derivative of the Integrated DEM dataset. A –scatterplot between MBES and Integrated DEM; B – scatterplot between LiDAR and Integrated DEM; C – slope derivative limited to 0-15 degrees; D-F – close-up examples of slope feature located in panel C.

The merging of spatial datasets often leads to issues at the intersection or overlap of different datasets. Since these are often not visible in bathymetry, we derived the slope feature of bathymetry to monitor subtle changes that may occur in these locations (Figure 8C). The slope feature reveals varying degrees of smoothness between bathymetric areas of different origins. Generally, areas derived from MBES appear to have a smoother relief compared to areas of LiDAR origin. A more detailed examination of further close-up examples (Figure 8D-F) reveals a smooth transition between the datasets. Based on these results, we can confidently assert that the integrated dataset is of high quality and can be utilized as input for further analysis and interpretation of the Puck Lagoon.

## Acknowledgements

The authors would like to acknowledge the crew of research vessels IMOROS 2 and IMOROS 3 for their competent assistance during the surveys. This research was funded in whole or in part by National Science Centre, Poland [Grant number: 2021/40/C/ST10/00240]. The contribution of Dr. Panagiotis Agrafiotis is part of MagicBathy project funded by the European Union's HORIZON Europe research and innovation programme under the Marie Skłodowska-Curie grant agreement No 101063294.

## Author contributions
Łukasz Janowski conceived the idea, acquired funding for the research, organised and supervised the team, coordinated—and personally contributed to—the acquisition and processing of MBES data and to the whole database. Łukasz Janowski also wrote and reviewed the original manuscript. Dimitrios Skarlatos mentored the team, processed airborne photogrammetry datasets and formulated ideas for airborne photogrammetry data collection. Panagiotis Agrafiotis wrote the airborne photogrammetry and SDB sections of the paper, provided software for refraction correction, personally contributed to data processing and technical validation of airborne photographs and SDB. Paweł Tysiąc authored the bathymetric LiDAR sections of the paper, contributed personally to the curation of bathymetric LiDAR data, and validated the dataset through formal analysis. Additionally, he originated the concept of bathymetry point cloud processing for seabed extraction. Andrzej Pydyn, Mateusz Popek and Anna M. Kotarba-Morley wrote the background and summary part of the paper and contributed information about underwater cultural heritage. Gottfried Mandlburger investigated and validated the airborne LiDAR dataset. Łukasz Gajewski and Mateusz Kołakowski processed and validated the MBES bathymetry dataset. Alexandra Papadaki developed methodology and performed data processing of SDB. Juliusz Gajewski provided project administration and operational opportunities to perform MBES surveys.

## Code Availability
No custom code was generated for this work.

## Competing interests
The authors declare no conflicts of interest.